\documentclass[runningheads]{llncs}

\usepackage[T1]{fontenc}
\def\doi#1{\href{https://doi.org/\detokenize{#1}}{\url{https://doi.org/\detokenize{#1}}}}
\usepackage{graphicx}
\usepackage{hyperref}
\usepackage{listings}
\usepackage{subfigure}
\usepackage{soul}
\usepackage{caption}
\usepackage{enumerate}
\usepackage{amsmath}
\usepackage{amsfonts}
\usepackage{xcolor}
\usepackage{wrapfig}
\usepackage{booktabs}
\usepackage{multirow}
\usepackage{amsmath}

\DeclareMathOperator*{\argmin}{arg\,min}

\newcommand{\adabracket}[1]{\left(#1\right)}

\newcommand{\entropy}[1]{\widehat{\mathcal{H}}\adabracket{#1}}
\newcommand{\approxEntropy}[1]{\widehat{\mathcal{H}}\adabracket{#1}}
\newcommand{\vx}{\boldsymbol{x}}
\newcommand{\vz}{\boldsymbol{z}}
\newcommand{\prob}[2]{P(#1|#2)}
\newcommand{\X}{\mathcal{X}}
\newcommand{\Y}{\mathcal{Y}}
\newcommand{\sumk}[1]{\sum_{k=#1}^K}
\newcommand{\mynorm}[2]{\left|\left|#1\right|\right|_{#2}}
\newcommand*\samethanks[1][\value{footnote}]{\footnotemark[#1]}

\begin{document}
\title{Automated Cancer Subtyping via Vector Quantization Mutual Information Maximization
}

\titlerunning{Abbreviated paper title}
\author{Zheng Chen\thanks{indicates joint first authors.}\inst{1} \and 
Lingwei Zhu\samethanks\inst{2}
\and 
Ziwei Yang\inst{3}
\and 
Takashi Matsubara\inst{1}
}
\authorrunning{F. Author et al.}
%
\institute{Osaka University, Japan \\
\and
University of Alberta, Canada \\
\and
Nara Institute of Science and Technology\\
\email{chen.zheng.bn1@gmail.com}
}

%
\maketitle              
\begin{abstract}
    Cancer subtyping is crucial for understanding the nature of tumors and providing suitable therapy. However, existing labelling methods are medically controversial, and have driven the process of subtyping away from teaching signals. Moreover, cancer genetic expression profiles are high-dimensional, scarce, and have complicated dependence, thereby posing a serious challenge to existing subtyping models for outputting sensible clustering. In this study, we propose a novel clustering method  for exploiting genetic expression profiles and distinguishing subtypes in an unsupervised manner. The proposed method adaptively learns categorical correspondence from latent representations of expression profiles to the subtypes output by the model. By maximizing the problem-agnostic mutual information between input expression profiles and output subtypes, our method can automatically decide a suitable number of subtypes. Through experiments, we demonstrate that our proposed method can refine existing controversial labels, and, by further medical analysis, this refinement is proven to have a high correlation with cancer survival rates.

\end{abstract}
\section{Introduction}\label{sec:intro}

Cancer is by far one of the deadliest epidemiological diseases known to humans: consider the breast cancer which is the most prevalent (incidence 47.8\% worldwide) and the most well-studied cancer in the world \cite{KDD21}, the 5-year mortality rate can still reach 13.6\% \cite{who}.
Its heterogeneity is considered as the crux of limiting the efficacy of targeted therapies and compromising treatment outcomes since some tumors that differ radically at the molecular level might exhibit highly resemblant morphological appearance \cite{moresubtype1}.
Increasing evidence from modern transcriptomic studies has supported the assumption that each specific cancer is composed of multiple categories (known as cancer subtypes) \cite{cancerhassubtype,KDDexploration}.
Reliably identifying cancer subtypes can significantly facilitate the prognosis and personalized treatment \cite{bioinformatics2021}.
However, currently there is a fierce debate in the cancer community: given transcriptomic data of one cancer, authoritative resources put that there might be different number of subtypes from distinct viewpoints, that is, the fiducial definition of the subtypes is constantly undergoing calibration \cite{PNAS}, suggesting for the majority of cancers the ground-truth labeling remains partially unavailable and awaits better definition. 

In the data science community, the lack of ground truth for the cancer data can be addressed as a clustering problem \cite{Grandvalet2004-semisupervisedByEntropyMinimization}, in which the clusters give a hint on the underlying subtypes.
Such clustering methods rely crucially on the quality of the data and suitable representations.
Modern subtyping methods typically leverage molecular transcriptomic expression profiles (expression profiles in short) which consist of genetic and microRNA (miRNA) expressions that characterize the cancer properties \cite{de2010molecular,bioinformatics2021}.
However, several dilemmas exist in the way of fully exploiting the power of expression profiles: 
\begin{itemize}
    \item \emph{High-dimensionality}: the expression profiles are typically of $>60,000$ dimensions; even after typical preprocessing the dimension can still be $> 10,000$.
    \item \emph{Scarcity}: cancer data are scarce and costly. Even for the most well-studied breast cancer, the largest public available dataset consists of expression profiles from around only 1500 subjects \cite{TCGA};
    \item \emph{Dependence}: expression profiles have complicated dependence: a specific expression might be under joint control of several genes, and sometimes such the joint regulation can be circular, forming the well-known gene regulation network \cite{KDD18}.
\end{itemize}
To extract information from the inherently high-dimensional expression profiles for tractable grouping \cite{PKDD18}, traditional methods preprocess the data via variants of principal components analysis (PCA) or least absolute shrinkage and selection operator (LASSO) \cite{pca_kmeans_2} for reducing the dimensionality of the data.
However, expression profiles with such complicated dependence have already been shown to not perform well with PCA and LASSO \cite{PKDD14}, since many seemingly less salient features can play an important role in the gene regulation network.
Motivated by the resurgence of deep learning techniques, recently the community has seen promising applications leveraging deep autoencoders (AEs) or variational AEs (VAEs) for compressing the data into a lower-dimensional latent space that models the underlying genetic regulation \cite{cancerhassubtype}.
However, VAEs with powerful autoregressive decoders often ignore the latent spaces \cite{Oord2017-VQVAE-neuralDiscreteRepresentation,chen2017-variationalLossyEncoder},  which runs the risk of overfitting \cite{Springenberg2015-un-supervisedCategoricalGAN}.
Furthermore, the latent representation is assumed to be continuous variables (usually Gaussian) \cite{Kingma2014-VaritionalBayesAutoEncoder,guoyikebriefing}, which is at odds with the inherently categorical cancer subtypes \cite{brca_s1}.
As a result, those subtyping models might have poor performance as well as generalization ability.

Aside from feature extraction, another issue concerns the grouping process itself.
Given extracted features from the expression profiles, the above-mentioned methods usually apply similarity-based clustering algorithms such as K-means for subsequent grouping.
However, such methods require strong assumptions on the data and are sensitive to representations \cite{PNAS2005-informationBasedClustering}: one will have to define a similarity metric for the data (often Euclidean) and find appropriate transformations (such as logarithm transform) as informative features.
Unsuitable choices of the metric and transformation can greatly degrade the model performance. 
Recently, mutual information has been gaining huge popularity in deep representation learning as a replacement for similarity metrics \cite{Hu2017-discreteSelfAugmentingRIM,Boudiaf2020-transductiveInfoMaxFewShot}: it is the unique measure of relatedness between a pair of variables invariant to invertible transformations of the data, hence one does not need to find a \emph{right} representation \cite{Liang2020-reallyAccessSourceData}.
Better yet, if two genes share more than one bit of information, then the underlying mechanism must be more subtle than just \emph{on} and \emph{off}.
Such subtlety and more general dependence can be captured by the mutual information \cite{PNAS2005-informationBasedClustering}.

In this paper, we propose a novel, generally applicable clustering method that is capable of fully exploiting the expression profiles and outputting sensible cancer subtyping solutions.
Besides tackling the above-mentioned problems in a unified and consistent manner, the proposed method has an intriguing property of automatically adjusting the number of groups thanks to its special architecture, which stands as a sheer contrast to prior methods that predetermine the number of groups by domain knowledge.
Before introducing the proposed architecture in Section \ref{sec:method}, we summarize our contributions as follows: 
\begin{itemize}
    \item (Algorithmic) Inspired by recent work, we propose a novel clustering method vector quantization regularized information maximization (VQ-RIM) for cancer subtyping. VQ-RIM maximizes mutual information in the categorical VQ-VAE model, which results in a combination of VAE reconstruction loss and mutual information loss. (Section \ref{sec:method_proposed})
    \item (Effective)  We compare the clustering results of VQ-RIM against existing ground truth labels (together with controversial labels from the entirety of labels) on different cancer datasets and find that VQ-RIM concords well with the ground truth, which verifies the correctness of VQ-RIM. (Section \ref{sec:groundtruth} and \ref{sec:label_flows})
    \item (Medical) {Extensive experiments on distinct cancers verify that VQ-RIM produces subtyping that consistently outperform the controversial labels in terms of enlarged separation of between-group life expectancies.
    The clearer separation suggests VQ-RIM is capable of better capturing the underlying characteristics of subtypes than controversial labels.
    }
    We believe such results are far-reaching in providing new insights into the unsettled debate on cancer subtyping.(Section \ref{sec:label_flows})
\end{itemize}

\section{Related Work}\label{sec:related}

\textbf{Feature Extraction for Subtyping. }
Building a model suitable for cancer subtyping is non-trivial as a result of the cancer data scarcity.
High dimensionality and data scarcity pose a great challenge to automated models for generating reliable clustering results \cite{guoyikebriefing}.
Conventionally, the problem is tackled by leveraging classic dimension reduction methods such as PCA \cite{pca_kmeans_2}.
However, since the progress of cancers is regulated by massive genes in a complicated manner (which themselves are under the control of miRNAs), brute-force dimension reduction might run the risk of removing informative features \cite{pca_cons}.
On the other hand, recently popular AE-based models \cite{cancerhassubtype,bioinformatics2021}, especially VAEs, construct the feature space by reconstructing the input through a multi-dimensional Gaussian posterior distribution in the latent space \cite{guoyikebriefing}.
The latent posterior learns to model the underlying causalities, which in the cancer subtyping context corresponds to modeling the relationship among expression profiles such as regulation or co-expression \cite{cancerhassubtype}.
Unfortunately, recent investigation has revealed that VAEs with powerful autoregressive decoders easily ignore the latent space.
As a result, the posterior could be either too simple to capture the causalities; or too complicated so the posterior distribution becomes brittle and at the risk of posterior collapse \cite{Alemi2018-fixBrokenELBO,Oord2017-VQVAE-neuralDiscreteRepresentation}.
Moreover, the Gaussian posterior is at odds with the inherently categorical cancer subtypes \cite{brca_s1}.

In this paper, we propose to leverage the categorical VQ-VAE to address the aforementioned issues: 
(i) VQ-VAE does not train its decoder, preventing the model from ignoring its latent feature space resulting from an over-powerful decoder; 
(ii) VQ-VAE learns categorical correspondence between input expression profiles, latent representations, and output subtypes, which theoretically suggests better capability of learning more useful features.
(iii) the categorical latent allows the proposed model to automatically set a suitable number of groups by plugging in mutual information maximization classifier, which is not available for the VAEs.\\
\textbf{Information Maximization for Subtyping. }
Cancer subtyping is risk-sensitive since misspecification might incur an unsuitable treatment modality.
It is hence desired that the clustering should be \emph{as certain as possible for individual prediction, while keeping subtypes as separated as possible} \cite{Bridle1991-UnsupervisedClassifierMutualInfo,Grandvalet2004-semisupervisedByEntropyMinimization}.
Further, to allow for subsequent analysis and further investigation of medical experts, it is desired that the method should output probabilistic prediction for each subject.
In short, we might summarize the requirements for the subtyping decision boundaries as
(i) should not be overly complicated; 
(ii) should not be located at where subjects are densely populated;
(iii) should output probabilistic predictions.
These requirements can be formalized via the information-theoretic objective as maximizing the mutual information between the input expression profiles and the output subtypes \cite{Gomes2010-RIM,Tschannen2020-OnInfoMaxRepresentation}.
Such objective is problem-agnostic, transformation-invariant, and unique for measuring the relationship between pairs of variables.
Superior performance over knowledge-based heuristics has been shown by exploiting such an objective \cite{PNAS2005-informationBasedClustering}.

\section{Method}\label{sec:method}

\subsection{Problem Setting}\label{sec:method_clustering}

Let $\X$ be a dataset $\X = \{\vx_{1}, \dots, \vx_N\}$, where $\vx_{i} \in \mathbb{R}^{d}, 1 \leq i \leq N$ are $d$-dimensional vectors consisting of cancer expression profiles.
For a given $\vx$, our goal lies in determining a suitable cancer subtype $y \in \{1,2,\dots, K\}$ given $\vx$, where $K$ is not fixed beforehand and needs to be automatically determined. 
Numeric values such as $y=1,\dots, K$ do not bear any medical interpretation on their own and simply represent distinct representations due to the underlying data.
It is worth noting while a label set $\mathcal{Y}$ is available, it comprises a small subset of ground-truth labels $\mathcal{Y}_{gt} := \{y_{gt}\}$ that have been medically validated and a larger portion of \emph{controversial} labels $\mathcal{Y}_{c}:=\{y_c\}$, with $\mathcal{Y}_{gt} = \mathcal{Y}\textbackslash\mathcal{Y}_{c}$.
Our approach is to compare the clustering result $y$ of the proposed method against ground truth labels $y_{gt}$ to see if they agree well, as a first step of validation.
We then compare $y$ against controversial labels $y_{c}$ and conduct extensive experiments to verify that the proposed method achieves improvement upon the subtyping given by $y_c$.
Our goal is to unsupervisedly learn a discriminative classifier $D$ which outputs conditional probability $\prob{y}{\vx, D}$.
Naturally it is expected that $\sumk{1}\prob{y=k}{\vx, D} = 1$ and we would like $D$ to be probabilistic so the uncertainty associated with assigning data items can be quantitized.
Following \cite{Springenberg2015-un-supervisedCategoricalGAN}, we assume the marginal class distribution $\prob{y}{D}$ is close to the prior $P(y)$ for all $k$.
However, unlike prior work \cite{Gomes2010-RIM,Springenberg2015-un-supervisedCategoricalGAN} we do not assume the amount of examples per class in $\X$ is uniformly distributed due to the imbalance of subtypes in the data.

\subsection{Proposed Model}\label{sec:method_proposed}

\subsubsection{Information Maximization. }

Given expression profiles of subject $\vx$, the discriminator outputs a $K$-dimensional probability logit vector $D(\vx) \in \mathbb{R}^{K}$.
The probability of $\vx$ belonging to any of the $K$ subtypes is given by the softmax parametrization:
\begin{align*}
    \prob{y=k}{\vx, D} = \frac{e^{D_k(\vx)}}{\sumk{1}e^{D_k(\vx)}},
\end{align*}
where $D_k(\vx)$ denotes the $k$-th entry of the vector $D(\vx)$.
Let us drop the dependence on $D$ for uncluttered notation.
It is naturally desired that the individual prediction be as certain as possible, while the distance between the predicted subtypes as large as possible.
This consideration can be effectively reflected by the mutual information between the input expression profiles and the output prediction label.
Essentially, the mutual information can be decomposed into the following two terms:
\begin{align}
    \begin{split}
        \hat{I}(\vx, y)  \!:=\!  \underbrace{ -\sumk{1} \! P(y=k)\log \! P(y=k) }_{\entropy{P(y)}} +\, \alpha \underbrace{ \sum_{i=1}^{N}\frac{1}{N}\sumk{1} \prob{y=k}{\vx_i} \log\prob{y=k}{\vx_i} }_{-\approxEntropy{\prob{y}{ \X}}}.
    \end{split}
\end{align}
which are the marginal entropy of labels $\entropy{P(y=k)}$ and the conditional entropy $\approxEntropy{\prob{y}{\X}}$ approximated by $N$ Monte Carlo samples $\vx_i, i\in\{1,\dots, N\}$.
$\alpha$ is an adjustable parameter for weighting the contribution, setting $\alpha=1$ recovers the standard mutual information formulation \cite{Gomes2010-RIM}.
This formulation constitutes the \emph{regularized information maximization (RIM)} part of the proposed method.
The regularization effect can be seen from the following:
\begin{itemize}
    \item[$\boldsymbol{\cdot}$] Conditional entropy $\approxEntropy{\prob{y}{\X}}$ encourages confident prediction by minimizing uncertainty.
    It effectively captures the modeling principles that decision boundaries should not be located at dense population of data \cite{Grandvalet2004-semisupervisedByEntropyMinimization}. 
    \item[$\boldsymbol{\cdot}$] Marginal entropy $\entropy{P(y)}$ aims to separate the subtypes as far as possible. Intuitively, it attempts to keep the subtypes \emph{uniform}. Maximizing only $\approxEntropy{\prob{y}{\X}}$ tends to produce degenerate solutions by removing subtypes \cite{Boudiaf2020-transductiveInfoMaxFewShot,Gomes2010-RIM}, hence $\entropy{P(y)}$ serves as an effective regularization for ensuring nontrivial solutions.
\end{itemize}

\subsubsection{Categorical Latents Generative Feature Extraction. }
Recent studies have revealed that performing RIM alone is often insufficient for obtaining stable and sensible clustering solutions \cite{Boudiaf2020-transductiveInfoMaxFewShot,Liang2020-reallyAccessSourceData,Springenberg2015-un-supervisedCategoricalGAN}: Discriminative methods are prone to overfitting spurious correlations in the data, e.g., some entry $A$ in the expression profiles might appear to have direct control over certain other entries $B$.
The model might na\"ively conclude that the appearance of $B$ shows positive evidence of $A$.
However, such relationship is in general not true due to existence of complicated biological functional passways: Such pathways have complex (sometimes circular) dependence between $A$ and $B$ \cite{loop}.
Since discriminative methods model $\prob{y}{\vx}$ but not the data generation mechanism $P(\vx)$ (and the joint distribution $P(\vx, y)$) \cite{Grandvalet2004-semisupervisedByEntropyMinimization}, such dependence between genes and miRNAs might not be effectively captured by solely exploiting the discriminator, especially given the issues of data scarcity and high dimensionality.

A generative model that explicitly captures the characteristics in $P(\vx)$ is often introduced as a rescue for leveraging RIM-based methods \cite{Hu2017-discreteSelfAugmentingRIM,Lowe2019-putEndtoEndtoEnd,Springenberg2015-un-supervisedCategoricalGAN}.
Such methods highlight the use of VAEs for modeling the latent feature spaces underlying input $\X$: given input $\vx$, VAEs attempt to compress it to a lower-dimensional latent $\vz$, and reconstruct $\tilde{\vx}$ from $\vz$.
Recently there has been active research on leveraging VAEs for performing cancer subtyping \cite{guoyikebriefing,cancerhassubtype}.
However, existing literature leverage continuous latents (often Gaussian) for tractability, which is at odds with the inherently categorical cancer subtypes.
Furthermore, VAEs often ignore the latents which implies the extracted feature space is essential dismissed and again runs the risk of overfitting \cite{Alemi2018-fixBrokenELBO}.

We exploit the recent vector quantization variational auto-encoder (VQ-VAE) \cite{Oord2017-VQVAE-neuralDiscreteRepresentation} as the generative part of the proposed architecture.
The categorical latents of VQ-VAE are not only suitable for modeling inherently categorical cancer subtypes, but also avoids the above-mentioned latent ignoring problem \cite{Kingma2014-VaritionalBayesAutoEncoder}.
In VQ-VAE, the latent embedding space is defined as $\{e_i\} \in\mathbb{R}^{M\times l}$, where $M$ denotes the number of embedding vectors and hence a $M$-way categorical distribution.
$l < d$ is the dimension of each latent embedding vector $e_i, i\in\{1,\dots,M\}$.
VQ-VAE maps input $\vx$ to a latent variable $\vz$ via its encoder $z_e(\vx)$ by performing a nearest neighbor search among the embedding vectors $e_i$, and output a reconstructed vector $\tilde{\vx}$ via its decoder $\vz_q$.
VQ-VAE outputs a deterministic posterior distribution $q$ such that
\begin{align}
    q(\vz=k | \vx) = 
    \begin{cases}
        1, &\text{ if } k = \argmin_{j} \mynorm{z_e(\vx) - e_j }{2}^2 \\
        0, & \text{ otherwise }
    \end{cases}
    \label{eq:vqvae_posterior}
\end{align}
The decoder does not possess gradient and is trained by copying the gradients from the encoder.
The final output of the decoder is the log-posterior probability $\log\prob{\vx}{\vz_q}$ which is part of the reconstruction loss.

\subsubsection{Architecture and Optimization.}\label{sec:method_optimization}

\begin{figure}[t]
    \centering
    \includegraphics[width=\textwidth]{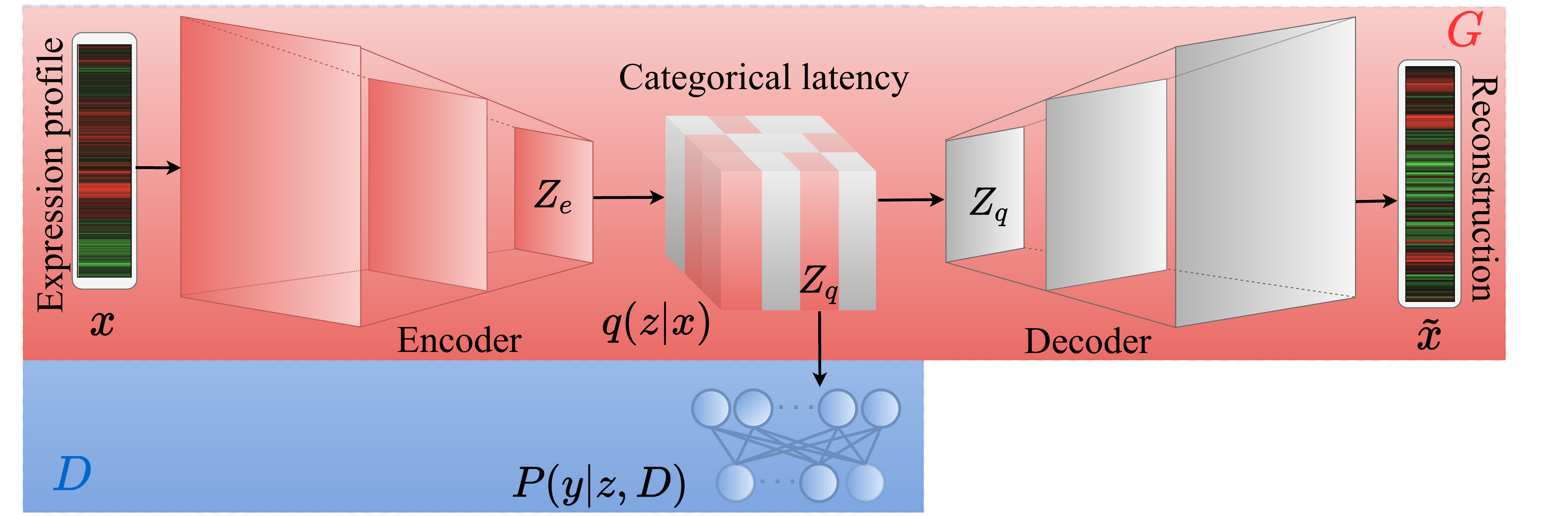}
    \caption{Overview of the proposed system.
    $D$ denotes the discriminator, $G$ denotes the generator.
    }
    \label{fig:overview}
\end{figure}

We propose a novel  model for clustering expression profiles as shown in Figure \ref{fig:overview}.
The model consists of a discriminator denoted as $D$ that maximizes the mutual information and a generator $G$ that aims to reconstruct the input via modeling a categorical underlying latent feature space spanned by $\{e_i\}$.
$D$ and $G$ are deeply coupled via the latent embeddings $\vz$, which is made possible through the fact the decoder of VQ-VAE does not possess gradients and hence the embedding space can be controlled by only the encoder and the discriminator.
In prior work, the generator is often architecturally independent from the discriminator and is only weakly related through loss functions \cite{Hu2017-discreteSelfAugmentingRIM,Liang2020-reallyAccessSourceData,Springenberg2015-un-supervisedCategoricalGAN}. 
Intuitively, one can consider the proposed model attempts to simultaneously minimize reconstruction loss as well as maximize the mutual information:

\begin{align}
        \mathcal{L} :=  \underbrace{ \entropy{P(y)} - {\approxEntropy{\prob{y}{\vz}}} - R(\lambda)}_{\mathcal{L}_{D}} + \underbrace{ \log\prob{\vx}{\vz_q} + \mynorm{\text{sg}[\vz_e] - e}{2} + \mynorm{\vz_e - \text{sg}[e]}{2} }_{\mathcal{L}_{G}}
    \label{eq:rim_objective}
\end{align}
where $\mathcal{L}_{D}, \mathcal{L}_{G}$ denote the discriminator loss and the generator loss, respectively.
$R(\lambda)$ is a possible regularizer that controls the weight growth, e.g. $R(\lambda) := \frac{\lambda}{2}||w^{T}w||^{2}_2$, where $w$ denotes the weight parameters of the model.
$\text{sg}[\cdot]$ denotes the stop gradient operator.

\subsubsection{Automatically Setting Number of Subtypes.}
The proposed model can automatically determine suitable number of subtypes by exploiting hidden information contained in the expression profiles which is not available to conventional methods such as K-means relying on prior knowledge.
The automatic subtyping is made possible via the deeply coupled latents and the discriminator: the multi-layer perceptron in the discriminator outputs the logarithm of posterior distribution $\log q(\vz | \vx)$.
However, by definition of Eq. (\ref{eq:vqvae_posterior}) the posterior is deterministic, which suggests $\log q(\vz | \vx)$ must either be $0$ or tend to $-\infty$.
The subsequent softmax layer hence outputs:
\begin{align}
    \prob{y=k}{\vz} = 
    \begin{cases}
        \frac{q(\vz = k | \vx)}{\sumk{1} q(\vz = k | \vx) },  & \text{ if } k = \argmin_{j} \mynorm{z_e(\vx) - e_j }{2}^2  \\
        0, & \text{ otherwise }
    \end{cases}
    \label{eq:vqrim_prob}
\end{align}
We can set $K$ to a sufficient large integer $\tilde{K}$ initially that covers the maximum possible number of subtypes.
Since the nearest neighbor lookup of VQ-VAE typically only updates a small number of embeddings $e_j$, by Eq. (\ref{eq:vqrim_prob}) we see for any unused $e_i, i\neq j$ the clustering probability is zero, which suggests the number of subtypes $K$ will finally narrow down to a much smaller number $K \ll \tilde{K}$.

\section{Experiments}\label{sec:results}

The expression profile data used in this study were collected from the world's largest cancer gene information database Genomic Data Commons (GDC) portal.
All of the used expression data were generated from cancer samples prior to treatment.

We utilized the expression profiles of three representative types of cancer for experiments:
\begin{itemize}
    \item Breast invasive carcinoma (BRCA): BRCA is the most prevalent cancer in the world. Its expression profiles were collected from the Illumina Hi-Seq platform and the Illumina GA platform.
    \item Brain lower grade glioma (LGG): the expression profiles were collected from the Illumina Hi-Seq platform.
    \item Glioblastoma multiforme (GBM): the expression profiles were collected from the Agilent array platform. Results on this dataset are deferred to the appendix.
\end{itemize}
These datasets consist of  continuous-valued expression profiles (feature length: 11327) of 639, 417 and 452 subjects, respectively.
Additional experimental results and hyperparameters can be seen in Appendix Section A available at \url{https://arxiv.org/abs/2206.10801}.


The experimental section is organized as follows:
we first compare the clustering results with the ground truth labels $\Y_{gt}$ in Section \ref{sec:groundtruth} to validate the proposed method.
We show in Section \ref{sec:controversial} that VQ-RIM consistenly re-assigns subjects to different subtypes and produces one more potential subtype with {enlarged separation in between-group life expectancies, which in turn suggests VQ-RIM is capable of better capturing the underlying characteristics of subtypes.}
Extensive ablation studies on both the categorical generator (VQ-VAE) and the information maximizing discriminator (RIM) are performed to validate the proposed architecture in Section \ref{sec:ablation}.
We believe the VQ-RIM subtyping result is far-reaching and can provide important new insights to the unsettled debate on cancer subtyping.






\subsection{Ground Truth Comparison}\label{sec:groundtruth}

\begin{wrapfigure}[18]{r}{0.5\textwidth}
    \vspace{-20pt}
    \begin{center}
      \includegraphics[width=0.45\columnwidth]{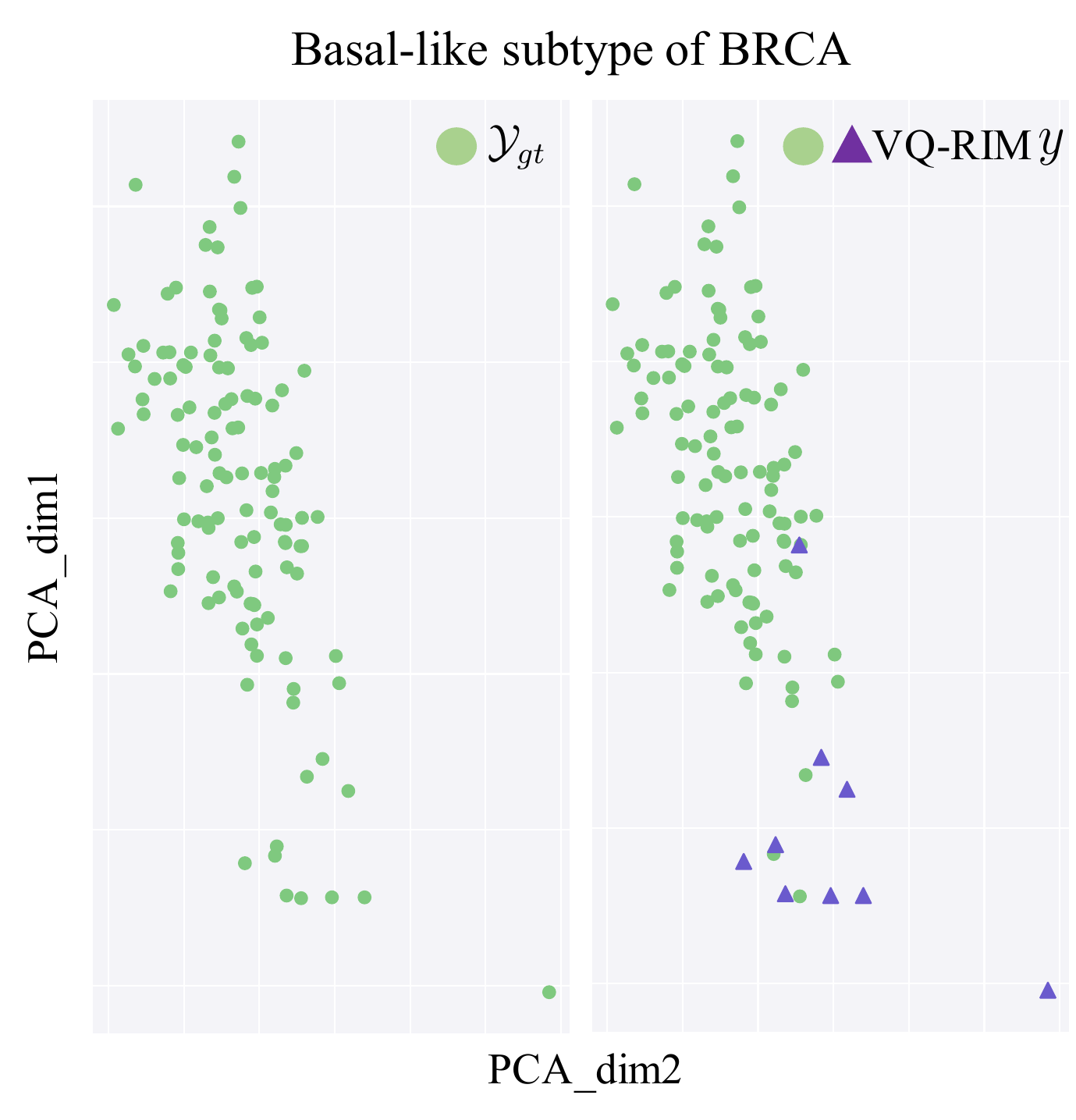}
    \end{center}
    \vspace{-20pt}
    \caption{Comparison between $\Y_{gt}$ and the VQ-RIM label $y$ on the Basal-like subtype of BRCA. 
    }
    \label{fig:groundtruth_compare}
\end{wrapfigure}
For validating the correctness of VQ-RIM, we show an example in Figure \ref{fig:groundtruth_compare}, i.e., the Basal-like cancer subtype of BRCA that has been well-studied and extensively validated by human experts and can be confidently subtyped, which can be exploited as the ground-truth labels $\Y_{gt}$.

However, other subtypes lack such well-verified labels and are regarded as the controversial labels $\Y_{c}$.
The left subfigure of Figure \ref{fig:groundtruth_compare} shows the two principal axes of Basal-like expression profiles after PCA.
The blue triangles in the right subfigure indicates the difference between $\Y_{gt}$ and the VQ-RIM result.
It can be seen that VQ-RIM agrees well with the ground truth.

\subsection{Controversial Label Comparison}\label{sec:controversial}

\subsubsection{Subtype Comparison. }\label{sec:subtypes_comparison}

\begin{figure}[t]
    \centering
    \includegraphics[width=0.93\columnwidth]{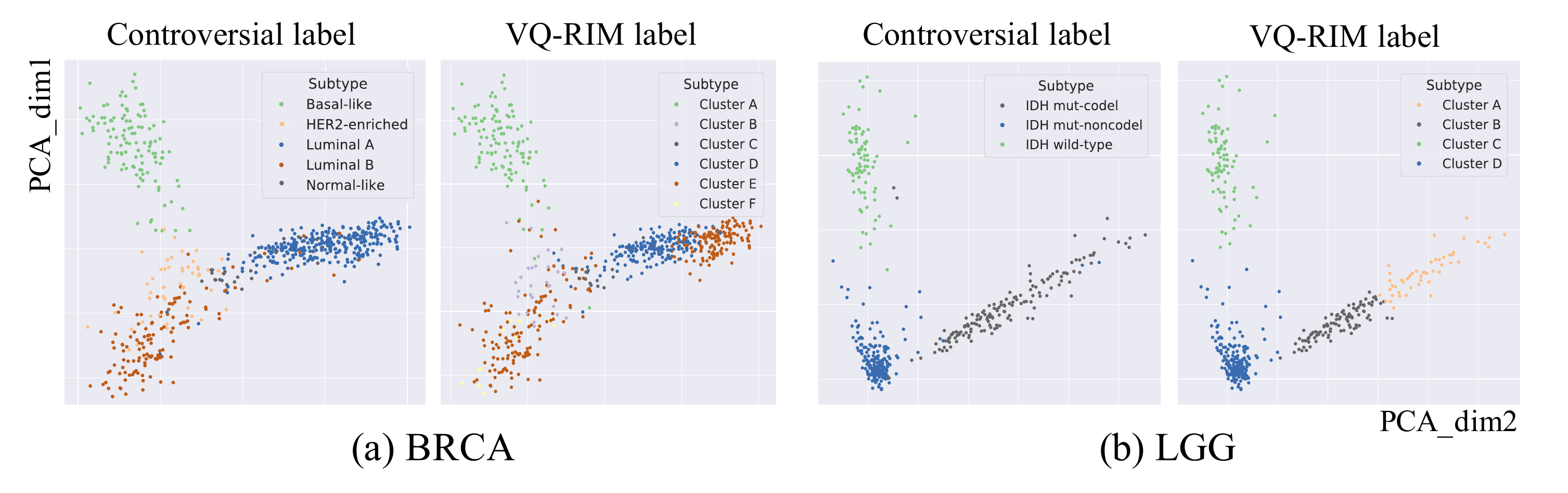}
    \caption{PCA visualization of the first two principal axes for BRCA and LGG. 
    }
    \label{fig:pca_lgg}
    \end{figure}

We compare existing controversial labels $\mathcal{Y}_c$ with the clustering results of VQ-RIM in Figure \ref{fig:pca_lgg}.
VQ-RIM output sensible decision boundaries that separated the data well and consistently produced one more subtype than $\mathcal{Y}_c$.
As confirmed in Section \ref{sec:groundtruth}, the Basal-like subtype concorded well with the VQ-RIM Cluster A.
On the other hand, other subtypes exhibited significant differences:
controversial labels seem to compactly fit into a fan-like shape in the two-dimensional visualization. 
This is owing to the human experts' heuristics in subtyping: intuitively, the similarity of tumors in the clinical variables such as morphological appearance often renders them being classified into an identical subtype.
However, cancer subtypes are the result of complicated causes on the molecular level.
Two main observations can be made from the BRCA VQ-RIM label:
(1) Luminal A was divided into three distinct clusters C,D,E. 
Cluster E now occupies the left and right wings of the fan which are separated by Cluster B and C;
(2) A new subtype Cluster F emerged from Luminal B, which was indistinguishable from Cluster E if na\"ively viewed from the visualization.
This counter-intuitive clustering result confirmed the complexity of cancer subtypes in expression profiles seldom admits simple representations as was done in the controversial labels.
A similar conclusion holds as well for other datasets such as LGG: 
IDH mut-codel was divided into two distinct subtypes (Cluster A, B), among which the new subtype Cluster A found by VQ-RIM occupied the right wing of IDH mut-codel.
In later subsections, the one more cluster and re-assignment of VQ-RIM are justified by analyzing the subtype population and from a medical point of view.
Due to page limit, we provide analysis focusing on BRCA only.

    \begin{figure}[t]
        \centering
        \includegraphics[width=0.85\columnwidth]{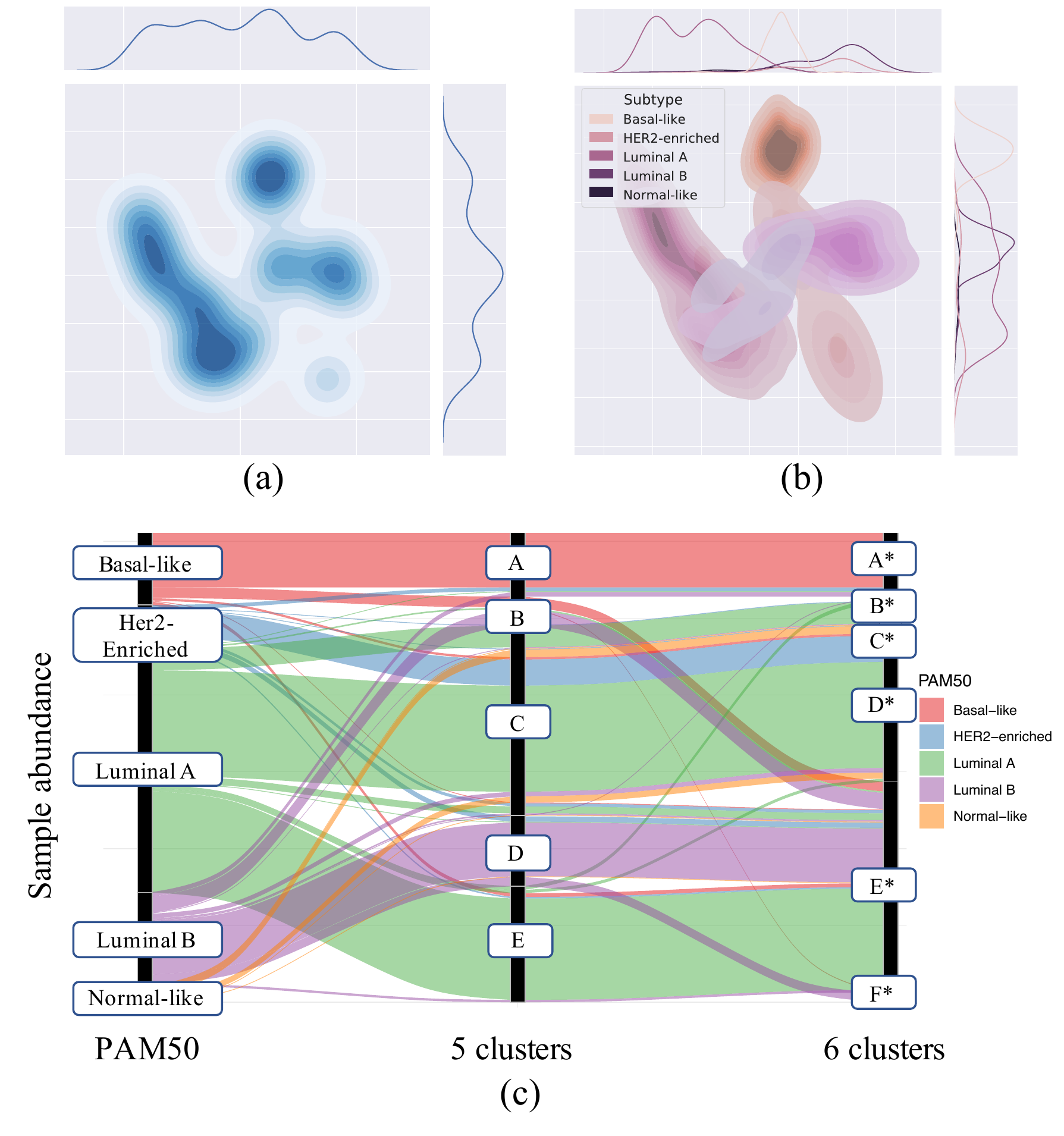}
        \caption{
        (a) t-SNE visualization of the sample distribution on BRCA. 
        (b) t-SNE of the samples with controversial labels.
        (c) label flows from the controversial labels (left) to VQ-RIM 5 subtypes (mid) and 6 subtypes (right).
        }
        \label{fig:density}
        \end{figure}

\subsubsection{Label Flows.}\label{sec:label_flows}

The controversial labels might run the risk of \emph{over-simplifying assignment} which refers to that in the regions overlapped with several distinct subtypes, controversial labels put all subjects into one of them without further identifying their sources. 
Such assignment can be illustrated by Figure \ref{fig:density}.
Here, Figure (\ref{fig:density}a) plots the sample distribution with darker colors indicating denser population of samples.
It is visible that the samples can be assigned to five clusters.
However, by injecting subtyping label information it is clear from Figure (\ref{fig:density}b) that in the lower left corner there existed strong overlaps of three different subtypes.
Controversial labels assigned them to a single subtype Luminal A.
VQ-RIM, on the other hand, was capable of separating those three subtypes. 
This separation can be seen from Figure (\ref{fig:density}c) which compares the two labeling when setting the number of VQ-RIM subtypes to 5 in accordance with controversial labels, or to 6 by setting $K$ to a sufficiently large value and automatically determines the suitable number of subtypes.
In either case, VQ-RIM consistently separated the Luminal A into three distinct subtypes: (B,C,E) in 5 subtypes case and (C\textsuperscript*, D\textsuperscript*, E\textsuperscript*) in the 6 subtypes case.
In the next subsection, we verify the effectiveness of such finer-grained subtyping by performing survival analysis.

\subsubsection{Medical evaluation. }\label{sec:medicalanalysis}

\begin{figure*}[t]
\centering
\includegraphics[width=0.93\linewidth]{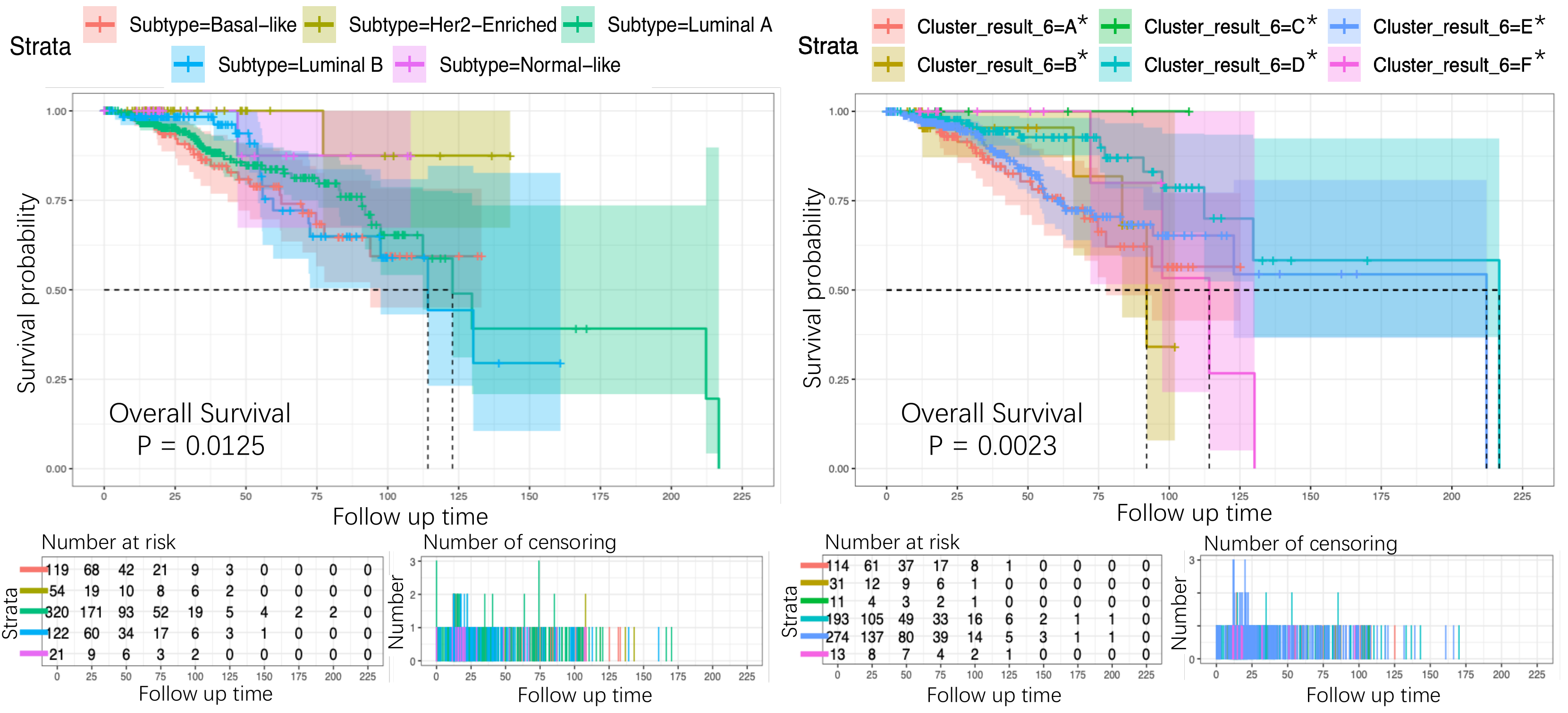}
\caption{Kaplan-Meier survival analysis within each identified subtype group (right) compared with original subtyping system (left) as a baseline.
The line in different colors represent patients from different subtypes.
P-value was calculated by Kaplan–Meier analysis with the log-rank test.}
\label{fig:km}
\end{figure*}

To demonstrate the clinical relevance of the identified subtypes, we perform subtype-specific survival analysis  by the Kaplan-Meier (KM) estimate.
{
KM estimate is one of the most frequently used statistical methods for survival analysis, which we use to complementarily validate the VQ-RIM labels from a clinical point of view}  \cite{km}.
KM compares survival probabilities in a given length of time between different sample groups.
The KM estimator is given by:
$\widehat{S}= \prod_{i:t{_{i}}<t}^{}\frac{n_{i}-d_{i}}{n_{i}}$,
where $n_{i}$ is the number of samples under observation at time $i$ and $d_{i}$ is the number of individuals dying at time $i$.
The survival analysis graph is plotted between estimated survival probabilities (on Y-axis) and the time passed after samples entry into the study (on X-axis), 
where the survival curve is drawn as a step function and falls only when a subject dies.

We can compare curves for different subtypes by examining gaps between the curves in horizontal or vertical direction. 
A vertical gap means that at a specific time point, 
samples belonging to one subtype had a greater fraction of surviving,
while a horizontal one means that it takes longer for these samples to experience a certain fraction of deaths.
The survival curves can also be compared statistically by testing the null hypothesis, i.e. there is no difference regarding survival situation among different groups,
which can be tested by classical methods like the log-rank test and the Cox proportional hazard test.

Figure \ref{fig:km} shows the KM survival analysis graph for BRCA samples, based on the PAM50 subtyping system and VQ-RIM subtypes.
Compared with the PAM50, the survival curves of VQ-RIM subtypes are more significantly separated.
Log-rank test also shows that there is significant difference in between-group survival with a smaller $p$-value of 0.0023 compared against the PAM50 ($\mathcal{Y}_c$).
{
Smaller $p$-values indicate better subtyping results.
}
We indicate the subtype-specific median survival time with dashed lines.
It is visible that VQ-RIM performed better in identifying subtypes with large median survival time differences.

\subsection{Ablation Study}\label{sec:ablation}





    \begin{figure}[t]
        \centering
        \includegraphics[width=0.86\columnwidth]{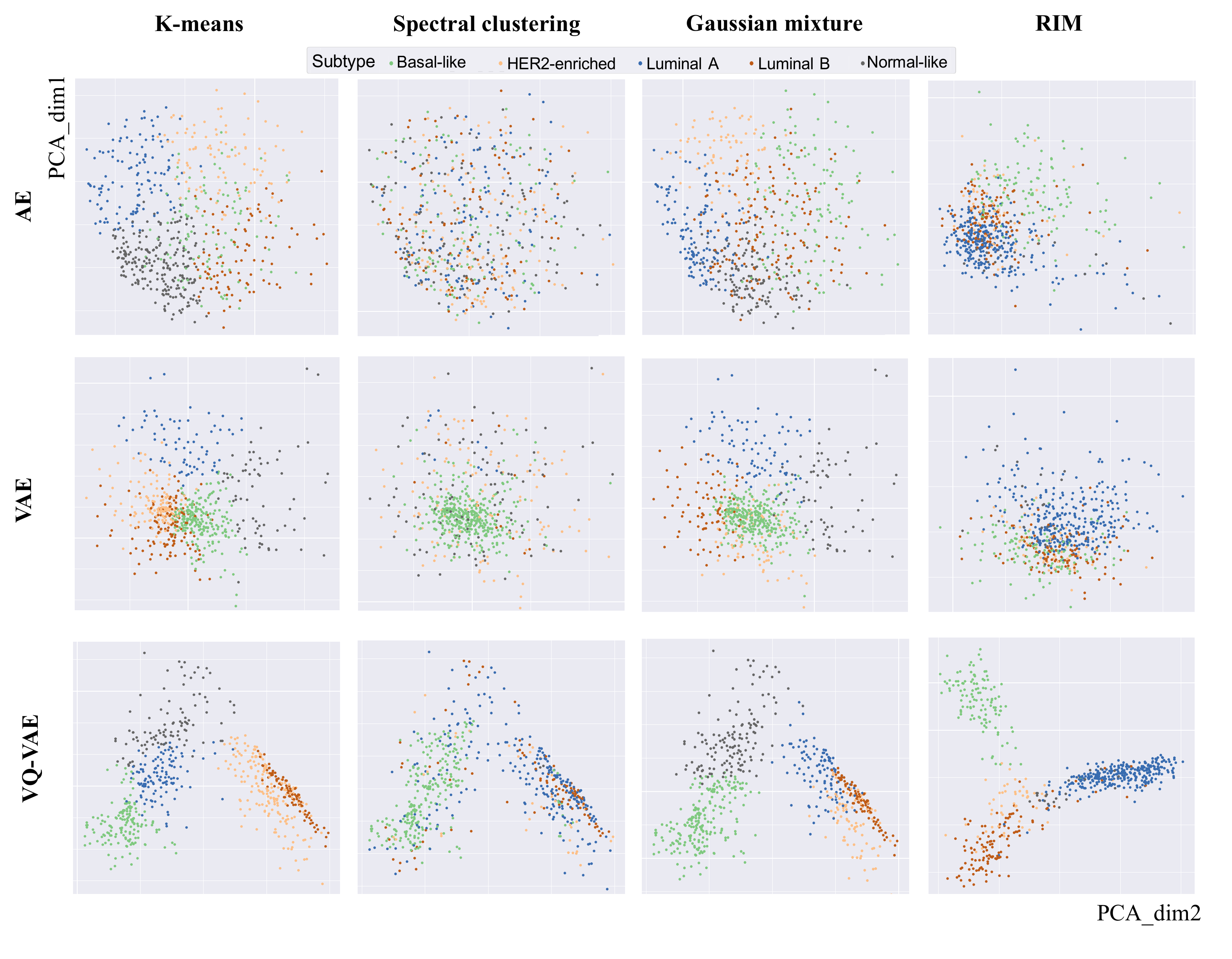}
        \caption{PCA visualization of the first two principal axes for BRCA using different clustering methods.
        The number of cluster number is determined by $\mathcal{Y}_c$ (PAM50).}
        \label{fig:ablation1}
        \end{figure}
    \begin{figure}[t]
        \centering
        \includegraphics[width=0.87\columnwidth]{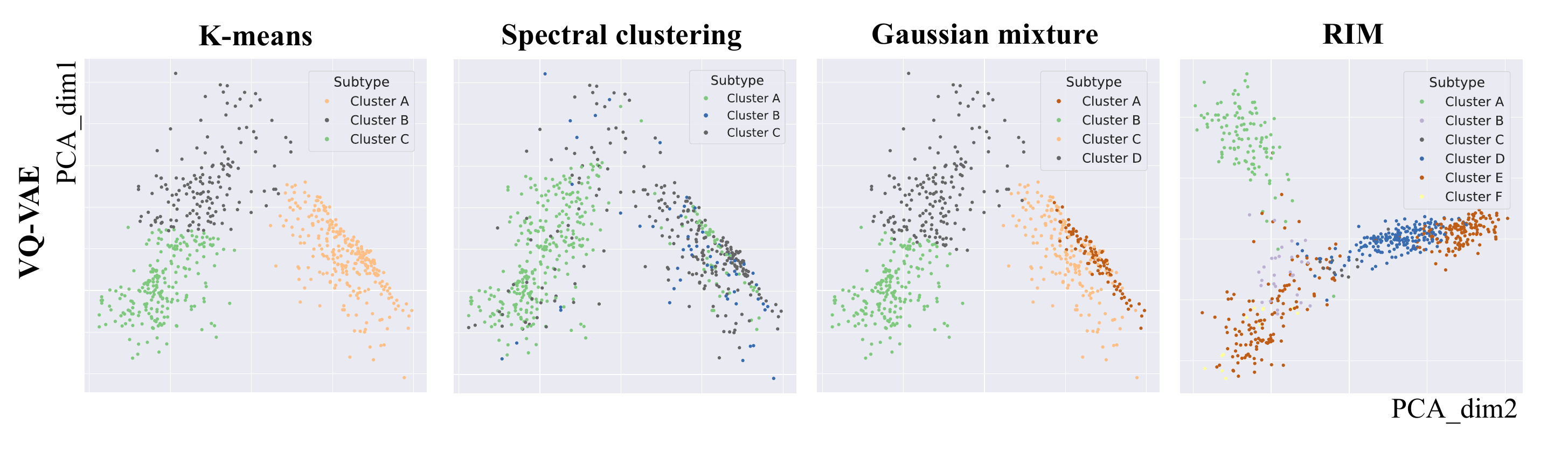}
        \caption{The number of clustering is determined by combining the results from the elbow method, not used for RIM.}
        \label{fig:ablation2}
        \end{figure}

In this section, we conduct comprehensive ablation experiments to further strength\-en the effectiveness of VQ-RIM.
Specifically, we validate the VQ part and RIM part respectively by comparing VQ-RIM against the following combinations:
\begin{itemize}
    \item we replace the VQ part with AE and VAEs with continuous latent which have been exploited for subtyping in \cite{guoyikebriefing,cancerhassubtype}. The expression profiles are compressed into continuous latent feature spaces for subsequent RIM clustering.
    \item we replace the RIM part with existing classic clustering algorithms such as K-Means, spectral clustering, and Gaussian mixture models \cite{pca_kmeans_2}. Categorical latent variables from VQ-VAE are fed into them for subtyping.
\end{itemize}
    Results of the ablation studies can be seen by inspecting Figure \ref{fig:ablation1} row-wise and column-wise, respectively.
    All methods applied $\mathcal{Y}_c$ (PAM50) for labeling.
    By inspecting the results row-wise, a first observation is that for all clustering methods AE and VAE tended to disperse data points.
    On the other hand, it can be seen from VQ-VAE clustering that subjects from distinct subtypes compactly located in lower-dimensional spaces.
    Column-wise inspection indicates that compared to other clustering methods, RIM tended to more cohesively aggregate the in-group points: 
    by contrast, it is visible from the VAE row that only Basal-like subjects were cohesively grouped.
    Among all compared methods, VQ-RIM stood out as the subjects were located in lower dimensional spaces from which clear decision boundaries emerged.
    The clear separation of VQ-RIM can be attributed to the underlying nearest neighbor search: such search essentially performed a preliminary grouping on the data, which greatly facilitated later separation.
    This observation is consistent with the recent finding of pseudo-labeling that explicitly performs K-means in creating preliminary clusters \cite{Liang2020-reallyAccessSourceData}.
    Besides the aforementioned qualitative evaluation, we also quantitatively measure in Table \ref{table:coefficients} the scores of all the clustering results in Figure \ref{fig:ablation1} by using the three well-accepted metrics: Normalized Mutual Information (NMI), Sihouette coefficient scores (Silhouette) and $p$-value of survival analysis \cite{kaufman2009finding}.

    However, in Section \ref{sec:controversial}, the labeling $\mathcal{Y}_c$ might not be the best medically even if the clustering result accords well with human intuition.
    In Figure \ref{fig:ablation2} we focus on VQ-VAE, and set the number of clusters for RIM to a sufficiently large value and let RIM automatically determines a suitable number of subtypes.
    We term this strategy automatic VQ-VAE in the last row of Table 
    \ref{table:coefficients}.
    
    For other clustering algorithms, the number of clusters is determined from the Silhouette coefficient scores and the elbow method \cite{kaufman2009finding}.
    It is visible that clustering algorithms other than RIM tended to reduce the number of subtypes for higher scores.
    By contrast, VQ-RIM produced one more subtype.
    This automatic VQ-RIM clustering was superior from a medical perspective since {it achieved greatest subtyping result as demonstrated by the smallest $p$-value of 0.0023.}
    Furthermore, algorithmically it is better than plain VQ-RIM as it achieved the highest scores of 0.63 and 0.54 of NMI and Silhouette among all ablation choices.

    \begin{table}
        \centering
        \caption{
        Metrics used for measuring the clustering results of Figure \ref{fig:ablation1} and \ref{fig:ablation2}. 
        The top three rows show the number of clustering determined by $\mathcal{Y}_c$ (PAM50), while the last row shows the number of clustering is automatically determined.
        }
        \vspace{5pt}
        \resizebox{.85\textwidth}{!}{
        \begin{tabular}{p{2cm}p{4cm}p{2cm}p{2cm}p{2cm}}
        \toprule[0.75pt]
                  \textbf{Generator}               &     \textbf{Discriminator}                & \textbf{NMI}\,$\uparrow$  & \textbf{Silhouette}\,$\uparrow$ & \textbf{$p$-value}\,$\downarrow$  \\ 
        \midrule[0.75pt]
        \multirow{4}{*}{AE +}     & K-Means             & 0.34 & 0.13       & 0.0861   \\
                                  & Spectral clustering & 0.01 & 0.01       & 0.1523   \\
                                  & Gaussian mixtures   & 0.34 & 0.11       & 0.0734   \\
                                  & RIM                 & 0.31 & 0.13       & 0.0834   \\ 
        \midrule
        \multirow{4}{*}{VAE +}    & K-Mmeans            & 0.29 & 0.17       & 0.0812   \\
                                  & Spectral clustering & 0.06 & 0.06       & 0.1382   \\
                                  & Gaussian mixtures   & 0.28 & 0.17       & 0.0899   \\
                                  & RIM                 & 0.33 & 0.22       & 0.0732   \\ 
        \midrule
        \multirow{4}{*}{VQ-VAE +} & K-Means             & 0.33 & 0.29       & 0.0154   \\
                                  & Spectral clustering & 0.05 & 0.04       & 0.0194   \\
                                  & Gaussian mixtures   & 0.44 & 0.24       & 0.0166   \\
                                  & RIM                 & 0.55 & 0.29       & 0.0042   \\ 
        \midrule
        \multirow{4}{*}{\begin{tabular}[c]{@{}l@{}}VQ-VAE +\\(Automatic)\end{tabular}} & K-Means             & 0.42 & 0.29       & 0.0145   \\
                                  & Spectral clustering & 0.06 & 0.05       & 0.0188   \\
                                  & Gaussian mixtures   & 0.51 & 0.32       & 0.0132   \\
                                  & RIM                 & \textbf{0.63} & \textbf{0.54}       & \textbf{0.0023}   \\
        \bottomrule
        \end{tabular}}
        \label{table:coefficients}
    \end{table}

\section{Discussion and Conclusion}\label{sec:conclusion}

In this paper we were concerned with the cancer subtyping problem that aimed to disclose the difference between subtypes within a specific cancer.
Existing literature is having an unsettled debate over the subtyping problem, with various definition and suitable number of subtypes put forward from distinct viewpoints.
Aiming to aid the medical experts by providing dependable reference for subtyping, we took a data-scientific standpoint and exploited genetic expression profiles of cancers without using the controversial labels current literature has imposed.
Such genetic expression profiles featured scarcity, high dimensionality, and complicated dependence which posed challenges for both physicians and data scientists.
To tackle these problems, we leveraged information-theoretic principles as well as recent categorical latent generative modeling techniques that helped in minimizing clustering confusion and maximizing interpretability.
The resultant novel model: Vector Quantization Regularized Information Maximization (VQ-RIM) can better reveal the intrinsic difference between cancer genetic expression profiles and based on which automatically decide a suitable number of subtypes.
The experiment on ground-truth BRCA cancer verified the correctness of VQ-RIM, while more extensive experiments on multiple authoritative datasets consisting of various cancers showed the difference between VQ-RIM results and the controversial labels.
By comprehensive analysis from both data scientific and medical views, we illustrated that the different subtyping result yielded by VQ-RIM consistently outperformed existing ones in terms of survival analysis, and contributed important new insights into the unsettled debate.

The future work consists of two interesting directions: 
(1) to further validate the effectiveness of VQ-RIM, comprehensive experiments on all available cancer datasets and comparison with their existing labeling might be necessary.
(2) the VQ-RIM architecture might not only work well with cancer data but also be generally applicable on radically different data such as images, voices that inherently exploit discrete nature of the data.

\section{Acknowledgement}

This work was supported by JST Mirai Program (JPMJMI20B8) and JST PRESTO (JPMJPR21C7), Japan.


%
%
%
\bibliographystyle{splncs04}
\bibliography{reference}
%





\clearpage
\appendix

\section{Parameter Settings and Training}\label{sec:param}

The details of the parameter settings are shown in Table \ref{tb:parameters}. 
To make the utmost of the model, a grid search of hyperparameters was implemented in this work to seek the best combination. 
Note that the optimal settings (values in Table \ref{tb:parameters}) were used both for all ablation studies. 
The experiments were conducted on a server with an NVIDIA GeForce RTX 3090Ti GPU.

\begin{table}
\caption{Parameter setting of experiments}
\centering
\begin{tabular}{l|c}
\toprule
~~~~~Model Parameter                          & ~~~Value~~~                                          \\ 
\midrule
~~~~~\#Embedding                          & ~~~64~~~                                                 \\
~~~~~Dimension of encoder~                 & ~~~512~~~                                                \\
~~~~~Dimension of embedding~               & ~~~64~~~                                                 \\
~~~~~Commitment cost                       & ~~~1~~~                                                  \\
~~~~~Dropout rate                          & ~~~0.5~~~                                                \\ \toprule
~~~~~Training settings                     & ~~~Value~~~                                              \\ 
\midrule
~~~~~\#Training epoch                      & ~~~200~~~                                                \\
~~~~~Batch size                            & ~~~32~~~                                                 \\
~~~~~Optimizer                             & ~~~AdamW~~~                                              \\
~~~~~Learning rate                         & ~~~$e \times 10^{-5}$~~~                        \\ \bottomrule
\end{tabular}
\label{tb:parameters}
\end{table}

\begin{figure*}
\centering
\includegraphics[width=0.95\linewidth]{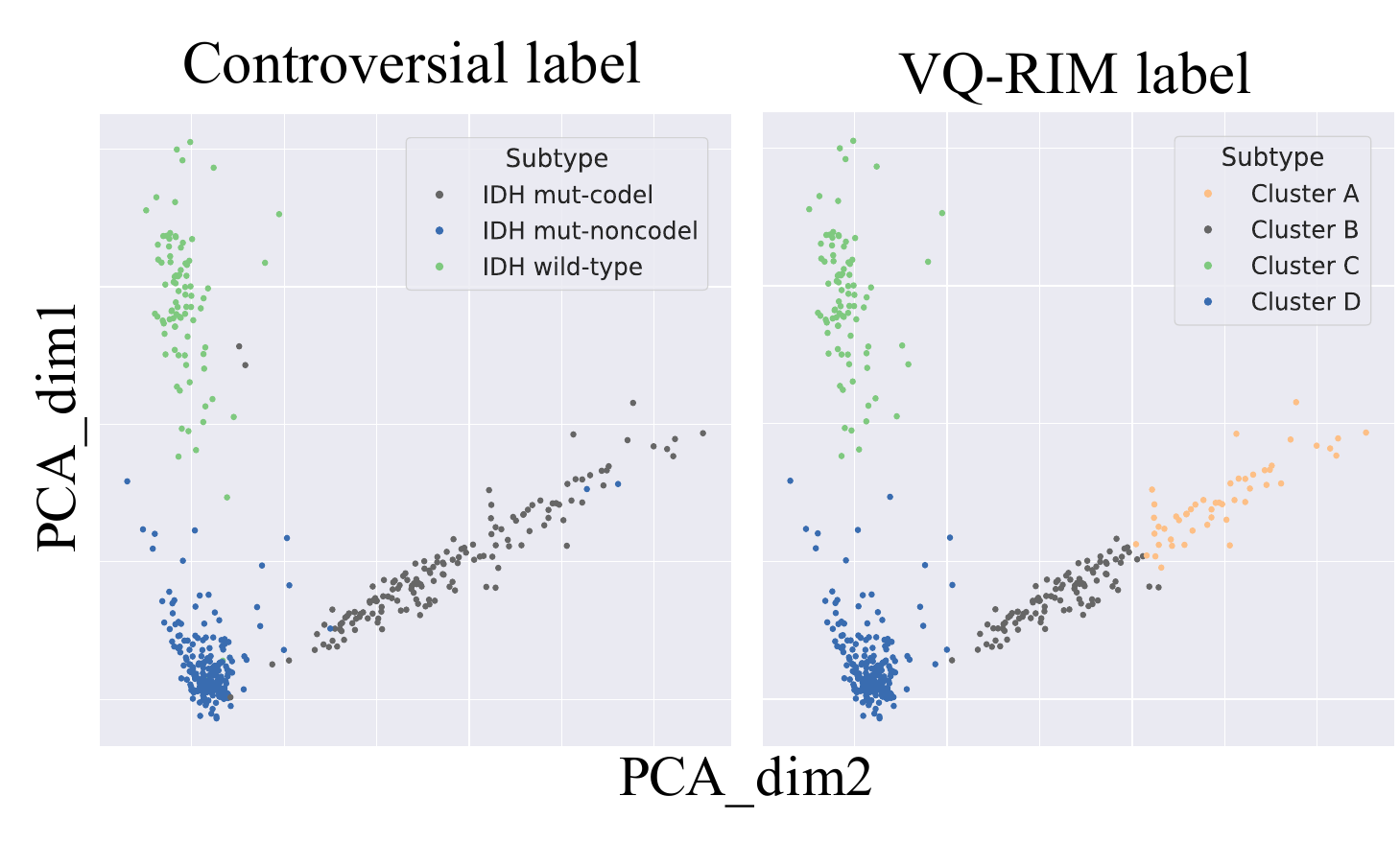}
\caption{PCA visualization of the first two principal axes for LGG.}
\label{fig_1}
\end{figure*}

This work is conducted a pre-training-to-fine-tuning strategy in the training process. 
Specifically, during the pre-training phase the all cancer sample from 4 cancers was used to initialize the model parameters of VQ-VAE. 
The $Adam$ with a biggish learning rate of $10^{-4}$ was utilized for the optimization.
In the fine-tuning phase, we implemented RIM on pre-trained VQ-VAE to conduct the subtyping tasks for different cancer datasets.
Here, the $AdamW$ optimizer was used to meticulously optimize the loss of subtyping.


\section{Complementary Results}

\begin{figure*}
\centering
\includegraphics[width=\linewidth]{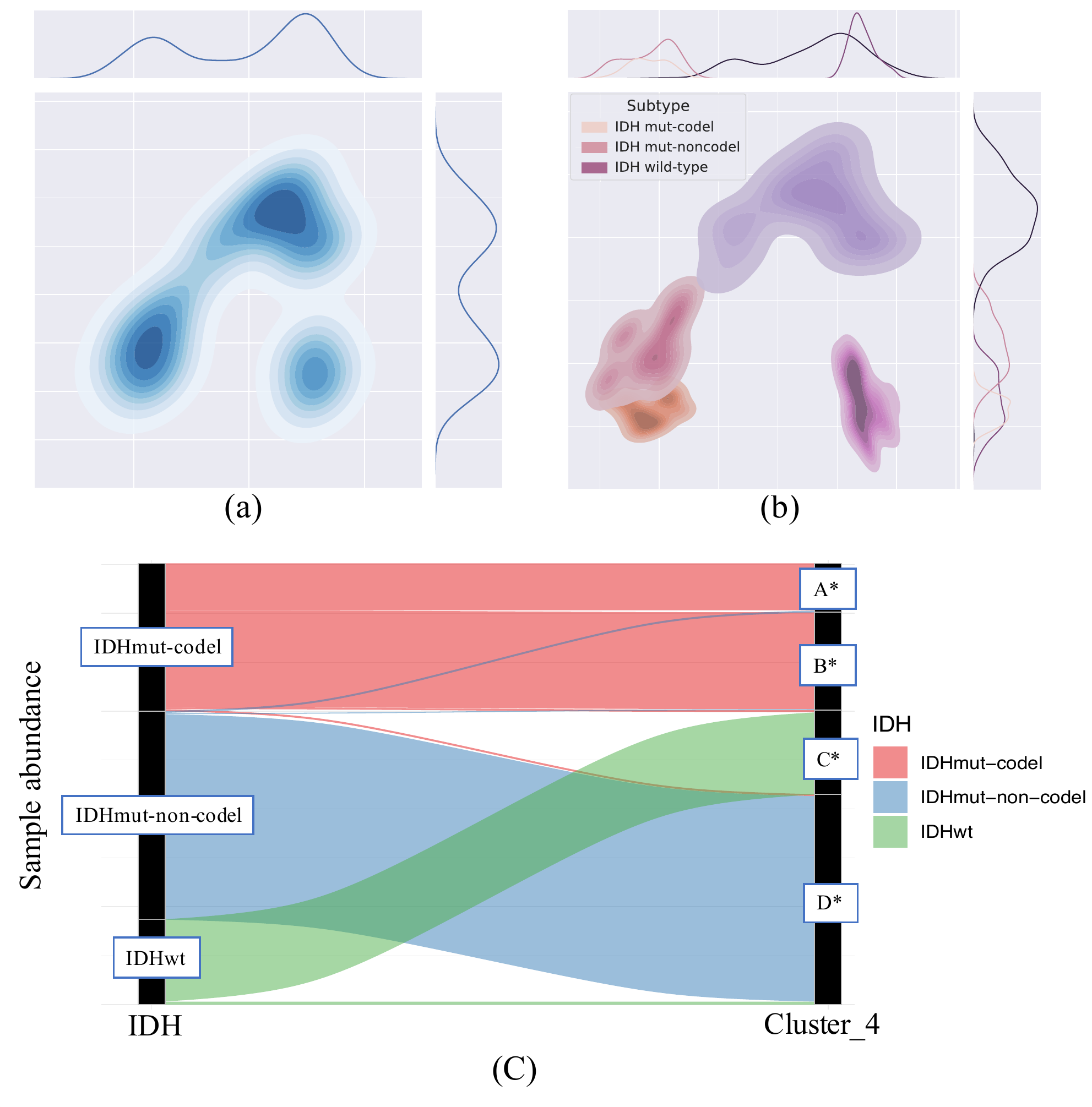}
\caption{(a) t-SNE visualization of the sample distribution on LGG. (b) t-SNE of the samples with controversial labels. (c) label flows from the controversial IDH cluster labels (left) to VQ-RIM 4 subtypes (right). }
\label{fig_2}
\end{figure*}

\begin{figure*}
\centering
\includegraphics[width=\linewidth]{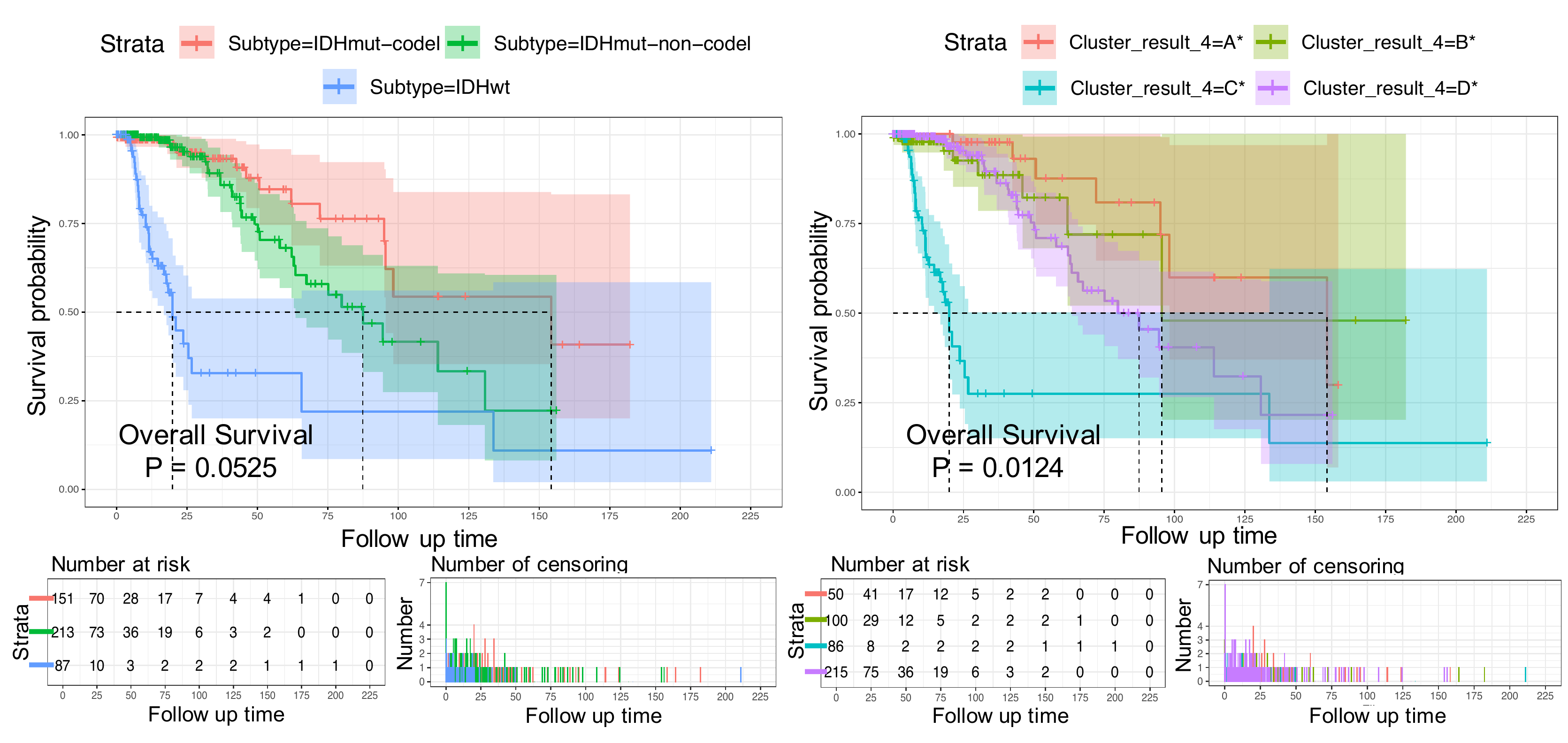}
\caption{Kaplan-Meier survival analysis within each identified subtype group (right) compared with original subtyping system (left) as a baseline.
The line in different colors represent patients from different subtypes.
P-value was calculated by Kaplan–Meier analysis with the log-rank test.}
\label{fig_3}
\end{figure*}

\begin{figure*}[t]
\centering
\includegraphics[width=\linewidth]{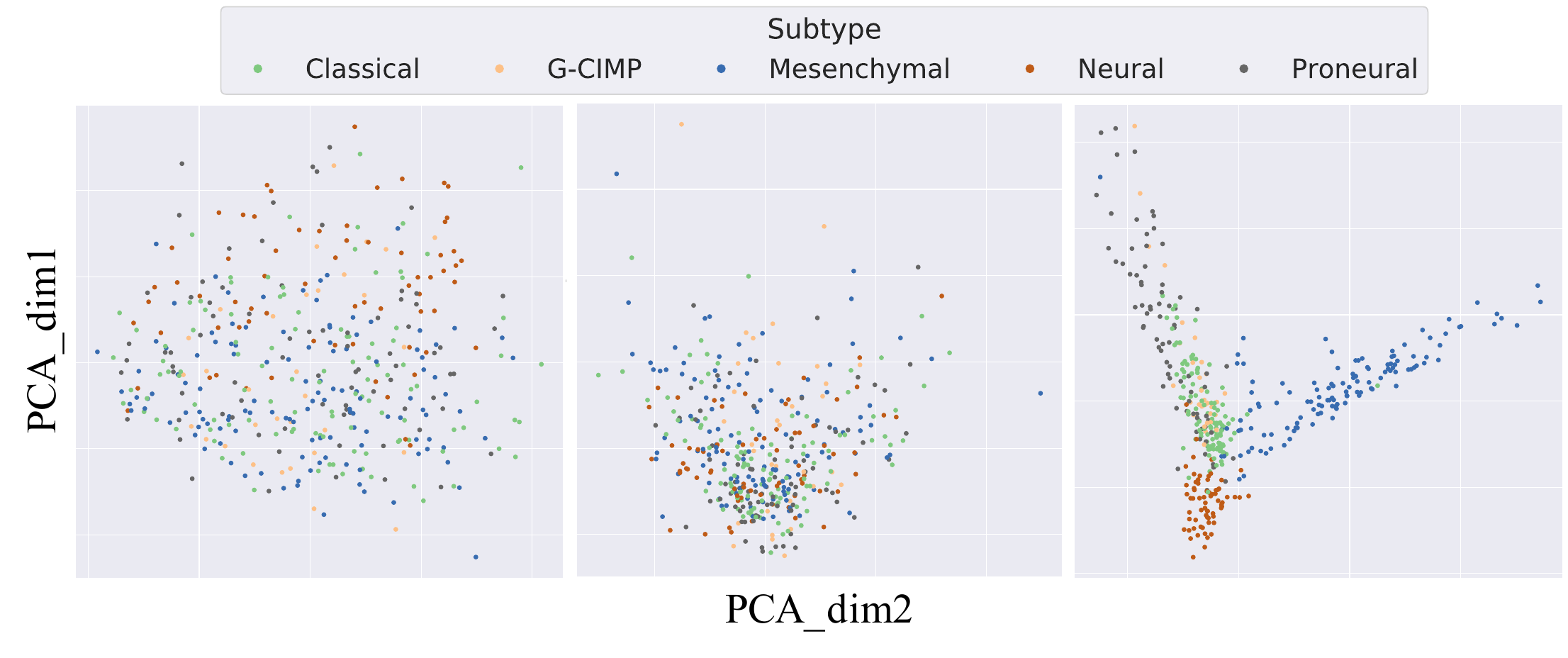}
\caption{PCA visualization of the first two principal axes for GBM.
It illuminates the proposed method has better clustering performance than others. }
\label{fig_4}
\end{figure*}

This section basically shows the results similar to Section 4 of the main paper.
Figure \ref{fig_1} is a reminiscent of Figure 3 in the main paper and shows a comparison between labeling $\mathcal{Y}_c$ and the number of subtypes determined automatically by VQ-RIM.
Figure \ref{fig_2} plots the density population density and label flows from $\mathcal{Y}_c$ and VQ-RIM which has one more subtype.
Figure \ref{fig_3} shows a KM analysis indicating the VQ-RIM result achieved clearer separation in the life expectancy with one more subtype.
The same conclusion holds for other datasets as well.
For simplicity we only show the clustering result for GBM in Figure \ref{fig_4}.

From Figure \ref{fig_1}, IDH mut-codel was divided into two distinct subtypes (Cluster A, B), among which the new subtype Cluster A found by VQ-RIM occupied the right wing of IDH mut-codel.
In later subsections, the one more cluster and re-assignment of VQ-RIM are justified by analyzing the subtype population and from a medical point of view.

Figure (\ref{fig_2}a) plots the sample distribution with darker colors indicating denser population of samples.
It is visible that the samples can be assigned to three clusters.
From Figure (\ref{fig_2}b) that we can find in the lower left corner there existed a overlaps of two different subtypes.
Controversial labels assigned them to a single subtype IDH mut-codel.
VQ-RIM was capable of separating those two subtypes. 
This separation can be seen from Figure (\ref{fig_2}c) which compares the two labeling when setting the number of VQ-RIM subtypes to 4 by setting $K$ to a sufficiently large value and automatically determines the suitable number of subtypes.
In either case, VQ-RIM consistently separated the IDH mut-codel into two distinct subtypes: (A* and B*) in 4 subtypes case.

Figure \ref{fig_3} shows the KM survival analysis graph for LGG samples, based on the IDH subtyping system and VQ-RIM subtypes.
Compared with the IDH, the survival curves of VQ-RIM subtypes are more significantly separated.
Log-rank test also shows that there is significant difference in between-group survival with a smaller $p$-value of 0.0124 compared against the IDH with 0.0525.

\end{document}